\documentclass[11pt]{article}
\usepackage[preprint]{acl}
\usepackage{times}
\usepackage{latexsym}
\usepackage{booktabs}
\usepackage{tabularx}
\usepackage{array}
\usepackage{amsmath}
\usepackage{multirow}
\usepackage{array}
\usepackage{graphicx}
\usepackage{placeins}
\usepackage{subcaption}
\usepackage{url}
\usepackage{float}
\usepackage{subcaption}
\usepackage[T1]{fontenc}
\usepackage[utf8]{inputenc}
\usepackage{microtype}
\usepackage{inconsolata}
\usepackage{amssymb}

\title{Same Probabilities, Different Advice: Gain-Loss Framing Effects in High-Stakes LLM Decision Support}

\author{
Yuyao Zhang$^{1}$, Yakun Cui$^{1}$, Han Zhu$^{1}$, Jinsong Lu$^{2}$, Yike Guo$^{1}$, Sirui Han$^{1\dagger}$\\[3pt]
{\normalfont\small
$^{1}$The Hong Kong University of Science and Technology, Hong Kong, China}\\
{\normalfont\small
$^{2}$Beihang University, Beijing, China}\\
{\normalfont\small siruihan@ust.hk}
}

\begin{document}
\maketitle

\begin{abstract}
Large language models (LLMs) are increasingly used to support consequential decisions, yet their reliability under equivalent representations of uncertainty remains unclear. Existing work has documented prompt sensitivity, cognitive biases, and confidence miscalibration, but it is still unknown whether high-stakes advice changes when the same probability is expressed through complementary gain and loss frames. We introduce \textsc{FrameBench-HPS}, a controlled benchmark of 2,880 prompts derived from 240 high-stakes scenarios across healthcare, finance, education/career, and legal/consumer-rights domains. Across seven LLMs, gain-loss framing changes the recommendation category in 32.4\% of complete groups, with most changes involving uncertainty rather than direct reversals. Human evaluation confirms corresponding shifts in perceived risk and directiveness. Robustness experiments show that framing effects persist under conversational paraphrases and exceed ordinary repeated-generation variability. Probability canonicalization substantially reduces framing sensitivity, identifying probability representation as a practical target for improving consistency in high-stakes LLM decision support.
\end{abstract}

\section{Introduction}

Large language models (LLMs) are increasingly capable of supporting complex decision-making tasks. Beyond retrieving information, current models can compare alternatives, synthesize evidence, reason over uncertain outcomes, and generate action-oriented recommendations. These capabilities have motivated applications in healthcare, finance, legal reasoning, and other consequential settings, where users may ask whether to pursue a treatment, make an investment, or take a legal action \cite{hu2026judge, he2026idea}. As LLMs become more capable and more frequently used as decision-support assistants, the reliability of their recommendations becomes increasingly important.

However, stronger task performance does not guarantee stable decision behavior. Recent studies show that LLM outputs can change under paraphrases, question-wording variations, and other meaning-preserving prompt reformulations \cite{takayama2026evaluating, alhetelah2026measuring}. Related work also reports framing and decision biases in applied settings, as well as unreliable abstention and confidence under uncertainty \cite{cocchieri2026llms, li2026demystifying}. 
These findings raise a practical concern for high-stakes advice: two users may express the same decision-relevant probability differently and receive different recommendations. Gain-loss framing provides a controlled test of this problem. For example, ``60\% chance of a favorable outcome'' and ``40\% chance of not achieving the favorable outcome'' describe the same binary probability structure but foreground different outcomes.
Despite recent efforts to improve prompt robustness \cite{tan2026basecal}, it remains unclear whether probability-equivalent framing systematically alters high-stakes recommendations beyond ordinary generation variability, and whether such sensitivity can be reduced through lightweight inference-time interventions.

\begin{figure}[t]
    \centering
    \includegraphics[width=\columnwidth]{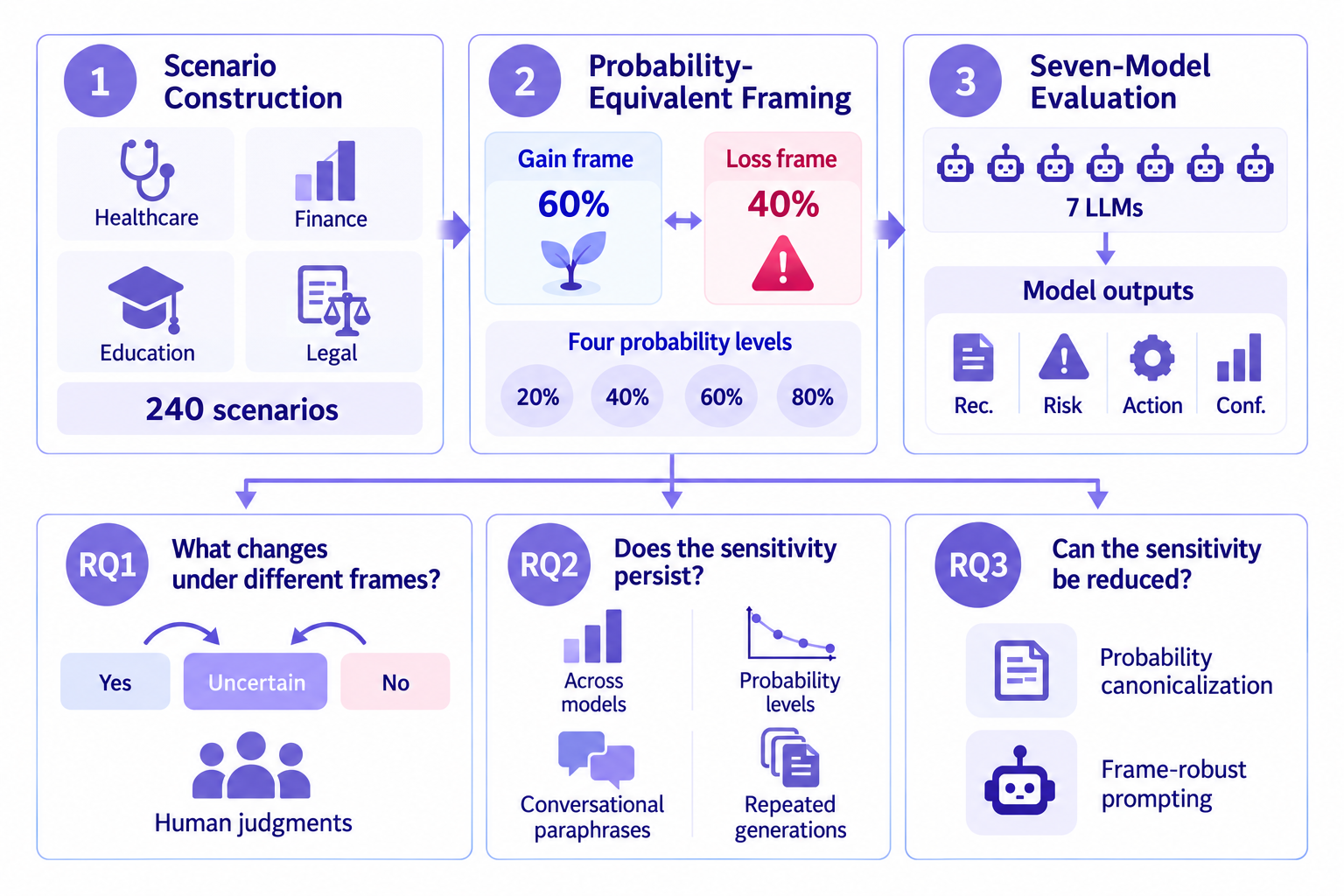}
    \caption{
    Overview of \textsc{FrameBench-HPS} and the evaluation framework.
    }
    \label{fig:framework}
\end{figure}

We address these gaps with \textsc{FrameBench-HPS}, a controlled benchmark of 2,880 prompts derived from 240 high-stakes scenarios across healthcare, finance, education/career, and legal/consumer-rights domains. Figure~\ref{fig:framework} summarizes the benchmark construction and the evaluation components corresponding to the three research questions. We study three questions:

\textbf{RQ1:} What forms of decision-support variation emerge under probability-equivalent gain-loss framing?

\textbf{RQ2:} How robust is framing sensitivity across models, probability levels, surface formulations, and repeated generations?

\textbf{RQ3:} To what extent can probability-representation interventions reduce framing sensitivity?

We evaluate seven LLMs on recommendation decisions and structured response measures, validate observable response properties through a three-annotator study, and conduct a three-model robustness evaluation using conversational paraphrases and repeated generations. We further test probability canonicalization and frame-robust prompting as inference-time interventions. Figure~\ref{fig:framework} summarizes the evaluation framework. Across the seven-model evaluation, gain-loss framing changes the recommendation category in 32.4\% of complete groups, while direct \texttt{yes}-\texttt{no} reversals remain rare. Under conversational prompting, cross-frame disagreement remains substantially higher than same-frame repeated-generation disagreement. Probability canonicalization reduces matched-repeat category changes to 6.7\% and nearly removes the mean framing-related risk shift.

Our contributions are:
\begin{itemize}
    \item We introduce \textsc{FrameBench-HPS}, a controlled benchmark for evaluating probability-equivalent framing sensitivity in high-stakes LLM decision support.
    \item We characterize framing sensitivity across seven LLMs and test its robustness using human judgments, conversational paraphrases, and repeated generations.
    \item We show that probability canonicalization substantially reduces framing-induced variation, identifying probability representation as a practical intervention point.
\end{itemize}

\section{Related Work}

\textbf{LLMs for high-stakes decision support.}
Recent work evaluates LLMs in consequential decision settings, including finance \cite{hu2026fin}, legal judgment \cite{hu2026judge}, and calibrated decision frameworks \cite{he2026idea}. In healthcare, models may still fail to abstain appropriately under uncertainty \cite{machcha2026knowing, cocchieri2026llms}, while confidence estimates can remain poorly calibrated \cite{lovering2025language, zhang2026confidence}. We study a complementary reliability property, which is whether advice remains stable under equivalent probability representations.

\textbf{Cognitive biases and framing in LLMs.}
Gain-loss framing is a well-established human bias in which equivalent descriptions can produce different judgments \cite{tversky1981framing,levin1998all}. LLM studies report related effects across broad bias benchmarks \cite{malberg2025comprehensive}, negotiation settings \cite{takenami2025does}, generated content \cite{alessa2025quantifying}, persona-conditioned decisions \cite{yoshikawa2026persona}, and risky-choice tasks \cite{ge2026mind}. DeFrame further studies semantically equivalent framing and framing-aware debiasing \cite{lim2026deframe}. Our work focuses on complementary probability framing in high-stakes advice and distinguishes uncertainty-mediated changes from direct recommendation reversals.

\textbf{Prompt sensitivity, robustness, and consistency.}
LLM behavior can vary across meaning-preserving prompt changes, including general reformulations \cite{liu2026understanding}, wording variations \cite{takayama2026evaluating}, paraphrases \cite{alhetelah2026measuring}, and non-adversarial prompt variants \cite{bhat2026all}. Recent work evaluates sensitivity and consistency directly \cite{errica-etal-2025-wrong, nalbandyan-etal-2025-score}, examines evaluation artifacts and human perceptions of consistency \cite{hua2025flaw, wu2025estimating}, and improves robustness through consistency training or uncertainty calibration \cite{hejabi2026flip, li2026demystifying}. \textsc{FrameBench-HPS} fixes the scenario and underlying probability, compares framing effects with repeated-generation variability, and evaluates inference-time mitigation without retraining.

\section{\textsc{FrameBench-HPS} Dataset}

\textsc{FrameBench-HPS} is designed to isolate changes in model decision support induced by probability-equivalent gain-loss framing.

\subsection{Scenario Abstraction}
We construct high-stakes probabilistic decision scenarios across four domains: healthcare, finance, education/career, and legal/consumer rights. Each scenario describes an actor considering a consequential action under uncertainty, such as adopting a treatment, making an investment, enrolling in a program, accepting a career opportunity, or pursuing a consumer remedy. 

Each base scenario is represented using a structured abstraction containing the domain, decision target, potential benefit, potential downside, probability-sensitive outcome, expected decision type, and high-stakes factor. The high-stakes factor records why the decision may have consequential personal outcomes. The abstraction separates scenario semantics from surface realization, allowing all framing variants of a scenario to be generated from the same underlying representation. 

The decision target specifies the action under consideration, while the benefit and downside fields encode consequences that remain fixed across framing conditions. The probability-sensitive outcome identifies the event whose likelihood is expressed differently under gain and loss framing. The expected decision type ensures that each scenario poses a concrete decision for which a recommendation can be elicited. All probability and framing variants derived from the same base scenario therefore share the same decision target, contextual facts, and outcome structure, enabling controlled comparisons under
semantics-preserving transformations
\cite{hong-etal-2025-consistencychecker}.

\subsection{Probability and Framing Conditions}
For each base scenario, we instantiate positive-outcome probabilities of 20\%, 40\%, 60\%, and 80\%. These levels span a broad range of uncertainty while avoiding near-deterministic endpoints. The benchmark targets framing sensitivity in uncertain decision regions. Gain-loss framing follows the established distinction between equivalent outcome descriptions that foreground favorable versus unfavorable outcomes \cite{tversky1981framing, levin1998all}.

Each probability level is paired with neutral, gain, and loss conditions. The neutral condition states the positive-outcome probability without explicit gain-oriented or loss-oriented lexical cues. Because it still refers to the favorable outcome, we treat comparisons involving this condition as secondary. The gain frame states the probability of the favorable outcome. The loss frame states the complementary probability that the favorable outcome will not occur. For example, ``80\% chance of success'' and ``20\% chance of not succeeding'' encode the same binary probability structure and differ in linguistic orientation.

\subsection{Prompt Generation and Quality Control}

We use an LLM-assisted, human-in-the-loop process to construct the benchmark, motivated by recent evidence that LLM-generated evaluation data require explicit validity and quality controls \cite{gill-etal-2025-lost}. GPT-5 generates an initial pool of 288 candidate decision scenarios, corresponding to 72 candidates for each of the four domains. We manually screen the candidates for decision clarity, realistic uncertainty, domain relevance, and the presence of a consequential action that can be expressed as a concrete recommendation. We then select and refine 60 base scenarios per domain, yielding 240 scenarios in total.

Fixed generation templates instantiate each retained base scenario at four positive-outcome probabilities (20\%, 40\%, 60\%, and 80\%) and under neutral, gain, and loss framing. For benchmark quality control, we first apply automated structural validation to the instantiated prompts. The checks verify required fields and formatting, domain balance, framing-group completeness, and exact probability complements within each gain-loss pair.

All structurally valid scenario groups then undergo human semantic review. Two annotators review all 240 base scenarios and their instantiated gain-loss formulations to verify that each pair refers to the same decision problem, expresses complementary probabilities of the same outcome, preserves the same decision target, contextual facts, and consequences, and introduces no additional asymmetric severity or evaluative cues beyond the intended framing manipulation. Any scenario that fails these criteria is revised, re-instantiated, and reviewed again. Revised instances are also rerun through the automated validation checks before inclusion in the final benchmark.
The resulting benchmark contains 2,880 prompts. Within each probability-equivalent group, the decision target, contextual information, and decision question remain fixed while the probability representation varies according to the framing condition. Representative scenario abstractions and instantiated prompts are provided in Appendix~\ref{app:dataset_examples}.

\section{Experimental Setup}

\noindent\textbf{Models.} We evaluate seven LLMs spanning multiple providers and model families: GPT-4o-mini \cite{openai2024gpt4omini}, GPT-4.1-mini \cite{openai2025gpt41}, GPT-5 \cite{openai2025gpt5}, Claude Sonnet 4.6 \cite{anthropic2026sonnet46}, Qwen3-VL-235B \cite{bai2025qwen3vl}, Gemini 3 Flash \cite{google2025gemini3flash}, and Gemini 2.5 Pro \cite{google2025gemini25pro}. Exact model/API identifiers and inference configurations are reported in Appendix~\ref{app:model_config}.

\noindent\textbf{Benchmark conditions.} We evaluate all 2,880 prompts in \textsc{FrameBench-HPS}. Each query is issued independently and does not expose other framing variants from the same probability-equivalent group. All models receive the same scenario format and structured-output schema. Of 20,160 expected model responses, 20,111 were recorded; the remaining 49 were not obtained after exhausting the predefined API retry procedure. Of the recorded responses, 20,079 contain verifiable final structured answers.

\noindent\textbf{Output schema and inference.}
The requested JSON output contains a categorical recommendation (\texttt{yes}, \texttt{no}, or \texttt{uncertain}), a 1-5 model-assigned risk score indicating how risky the proposed action is presented to be, a 1-5 actionability score indicating how strongly the response supports a concrete course of action, a 1-5 self-rated confidence score indicating the model's stated confidence in its recommendation, and a brief natural-language explanation. Appendix~\ref{app:prompt_template} provides the complete prompt and field definitions. We use temperature 0 for all models. The maximum output length is 1,500 tokens except for Qwen3-VL-235B, for which we use 8,192 tokens to accommodate its longer exposed reasoning trace. These values are truncation ceilings; all framing variants within a model use identical inference settings.

We apply explicit validity checks to all stored responses. Recommendations must belong to the predefined three-category set, scalar fields must contain integer values from 1 to 5, and an explanation must be present. Integral numeric strings are normalized without altering their value. For outputs containing visible reasoning, a structured answer is retained only when a complete JSON object appears after the closing \texttt{</think>} marker.

\noindent\textbf{Metrics.}
Following recent work that treats prompt sensitivity and consistency as reliability properties distinct from task accuracy \cite{errica-etal-2025-wrong,nalbandyan-etal-2025-score}, we quantify whether probability-equivalent gain and loss inputs produce different observable outputs. Let \(i\) index a scenario-probability-model combination and let \(y_i^G\) and \(y_i^L\) denote recommendations under gain and loss framing. The two inputs encode complementary outcomes of the same binary event:
\begin{equation}
P(O^{+})=p
\quad\Longleftrightarrow\quad
P(O^{-})=1-p.
\label{eq:prob_equiv}
\end{equation}

Recommendation-category change is
\begin{equation}
C_i=\mathbb{I}[y_i^{G}\neq y_i^{L}],
\label{eq:category_change}
\end{equation}
where
\(y_i^{f}\in\{\texttt{yes},\texttt{no},\texttt{uncertain}\}\).
Direct reversal is the stricter event
\begin{equation}
R_i=
\mathbb{I}
\left[
\{y_i^{G},y_i^{L}\}
=
\{\texttt{yes},\texttt{no}\}
\right].
\label{eq:direct_reversal}
\end{equation}

For structured scalar fields, let \(r_i^f\), \(a_i^f\), and \(c_i^f\) denote risk, actionability, and self-rated confidence under frame \(f\). We compute
\begin{equation}
\begin{aligned}
\Delta r_i &= r_i^{L}-r_i^{G}, &
\Delta a_i &= a_i^{G}-a_i^{L},\\
\Delta c_i^{\mathrm{abs}}
&=|c_i^{L}-c_i^{G}|, &
\Delta c_i^{\mathrm{signed}}
&=c_i^{L}-c_i^{G}.
\end{aligned}
\label{eq:scalar_shifts}
\end{equation}
Positive \(\Delta r\) indicates higher model-assigned risk under loss framing, while positive \(\Delta a\) indicates higher actionability under gain framing. Because the 1-5 fields are elicited without detailed scale anchors, we treat them as structured model outputs rather than validated psychological measures.

\noindent\textbf{Statistical analysis.}
For the seven-model evaluation, we report 95\% percentile bootstrap confidence intervals from 5,000 resamples of the 240 base scenarios. Each resampled scenario carries all associated probability levels and model observations, preserving within-scenario dependence. Scenario-level sign-flip tests provide inference for signed scalar differences, with Holm correction across risk, actionability, and signed self-rated confidence. All hypothesis tests are two-sided with a predefined significance threshold of \(\alpha=.01\).

\section{Results and Analysis}

Overall, we find that (1) gain-loss framing changes recommendation categories in 32.41\% of complete groups, with 98.1\% of these changes involving \texttt{uncertain}; (2) framing sensitivity persists under conversational reformulations and remains substantially larger than same-frame repeated-generation disagreement; and (3) probability canonicalization reduces matched-repeat category changes from 34.20\% to 6.68\%. We examine these findings below.

\subsection{RQ1 Framing Primarily Shifts Recommendations Through Uncertainty}
\label{sec:rq1}

\noindent\textbf{Setups.} We use 6,644 complete framing triplets for the primary analysis, where neutral, gain, and loss responses are all valid. To test whether the complete-triplet restriction affects the gain-loss comparison, we additionally analyze all 6,658 valid gain-loss pairs. We further conduct a human evaluation on 140 gain-loss response pairs selected through stratified random sampling, with 35 pairs per domain, 35 pairs per probability level, and 20 pairs per model, and all pairs drawn from distinct base scenarios. Three annotators independently rate both responses on 1-5 scales for \textit{risk presentation}, \textit{directiveness}, and \textit{expressed certainty}. Following prior work that distinguishes automatic consistency measures from human perceptions of model responses \cite{wu2025estimating}, annotators are blinded to model identity, framing condition, and the model-generated scalar fields.
Response-level scores are aggregated by the median across annotators. We report ordinal Krippendorff's $\alpha$, evaluate gain-loss differences using paired Wilcoxon signed-rank tests with Holm correction, and use Spearman's $\rho$ to compare human ratings with the corresponding structured model fields.

\noindent\textbf{Results.} Table~\ref{tab:results} provides a synopsis of the seven-model evaluation. Gain-loss framing changes the recommendation category in 2,153 of 6,644 complete groups, corresponding to 32.41\% (95\% CI 30.48-34.36\%). In contrast, only 42 groups directly reverse between \texttt{yes} and \texttt{no}, yielding a direct-reversal rate of 0.63\% (95\% CI 0.41-0.87\%).

\begin{table}[t]
\centering\scriptsize
\begin{tabular}{lr}
\toprule
Metric & Mean [95\% CI] \\
\midrule
Category change (\%) & 32.41 [30.48, 34.36] \\
Direct yes--no (\%) & 0.63 [0.41, 0.87] \\
Risk: loss$-$gain & 0.285 [0.259, 0.311] \\
Actionability field: gain$-$loss & -0.230 [-0.274, -0.187] \\
Self-rated confidence: absolute & 0.702 [0.674, 0.732] \\
Self-rated confidence: loss$-$gain & 0.352 [0.319, 0.387] \\
\bottomrule
\end{tabular}
\caption{
Aggregate gain--loss effects in the seven-model evaluation.
}
\label{tab:results}
\end{table}

Of the 2,153 category changes, 2,111, or 98.1\%, involve \texttt{uncertain}. The most frequent transition is from \texttt{uncertain} under gain framing to \texttt{no} under loss framing, observed in 1,301 groups, while another 681 groups change from \texttt{yes} under gain framing to \texttt{uncertain} under loss framing. These results indicate that framing primarily changes whether the model commits to a categorical recommendation, whereas direct \texttt{yes}-\texttt{no} reversals remain uncommon.

The structured fields exhibit corresponding differences. Loss framing increases model-assigned risk by 0.285 points on average (95\% CI 0.259-0.311), and all seven models show positive mean risk shifts. The gain-minus-loss actionability difference is $-0.230$ (95\% CI $-0.274$ to $-0.187$), indicating higher actionability under loss framing. The mean absolute self-rated confidence difference is 0.702 (95\% CI 0.674-0.732), with a signed loss-minus-gain difference of 0.352 (95\% CI 0.319-0.387). Scenario-level sign-flip tests support all three signed differences (Holm-adjusted $p=.0003$; Table~\ref{tab:signflip_tests}).

The human evaluation shows corresponding directional differences in perceived risk and directiveness. Risk-presentation ratings increase from 3.24 under gain framing to 3.51 under loss framing ($\Delta=0.27$, 95\% CI 0.17-0.38, $p_{\mathrm{Holm}}<.001$), while directiveness increases from 3.23 to 3.55 ($\Delta=0.32$, 95\% CI 0.14-0.49, $p_{\mathrm{Holm}}=.0015$). Expressed certainty shows no significant directional difference ($\Delta=0.08$, 95\% CI $-0.09$-0.24, $p_{\mathrm{Holm}}=.312$). Inter-annotator agreement is highest for risk presentation ($\alpha=.572$), followed by directiveness ($\alpha=.480$) and expressed certainty ($\alpha=.419$). Associations with the corresponding structured fields are $\rho=.704$, $.307$, and $.407$, respectively (Appendix~\ref{app:human_eval}).

The sensitivity analysis using all 6,658 valid gain-loss pairs produces a category-change rate of 32.44\% (95\% CI 30.51-34.39\%), compared with 32.41\% in the complete-triplet analysis. The direct-reversal rate remains 0.63\%, showing that the primary gain-loss result is unchanged when a valid neutral response is not required.

\subsection{RQ2 Framing Sensitivity Persists Beyond Fixed Prompt Templates}
\label{sec:rq2}

\noindent\textbf{Setups.} We first examine variation across the seven evaluated models and four probability levels. We then test whether the observed effect persists when fixed benchmark templates are replaced by conversational reformulations, following recent evaluations of semantically equivalent wording and paraphrase variation \cite{takayama2026evaluating, alhetelah2026measuring, kang-etal-2025-format}. For the robustness evaluation, we use GPT-4o-mini, Gemini 2.5 Pro, and Claude Sonnet 4.6. We randomly sample 10 base scenarios from each of the four domains, yielding 40 scenarios. Two scenarios per domain are used for development checks and eight per domain form the held-out test set, yielding 32 test scenarios. For each scenario, we manually construct one conversational paraphrase. Two authors review each paraphrase before inference to ensure that the decision target, factual constraints, numerical information, and gain-loss relation remain unchanged. Each condition is generated three times under the same inference configuration, allowing framing-related disagreement to be compared with repeated-generation variation of the same framed input \cite{arana-etal-2026-models}.
For repeated-generation analysis, matched-repeat disagreement compares gain and loss responses with the same repeat index. Same-frame disagreement measures pairwise variation among the three generations of the same framed input, while all-repeat cross-frame disagreement averages all nine gain-loss repeat combinations. Confidence intervals use 5,000 bootstrap resamples of the 32 held-out base scenarios.

\begin{figure*}[t]
    \centering
    \begin{subfigure}[t]{0.49\textwidth}
        \centering
        \includegraphics[width=\linewidth]{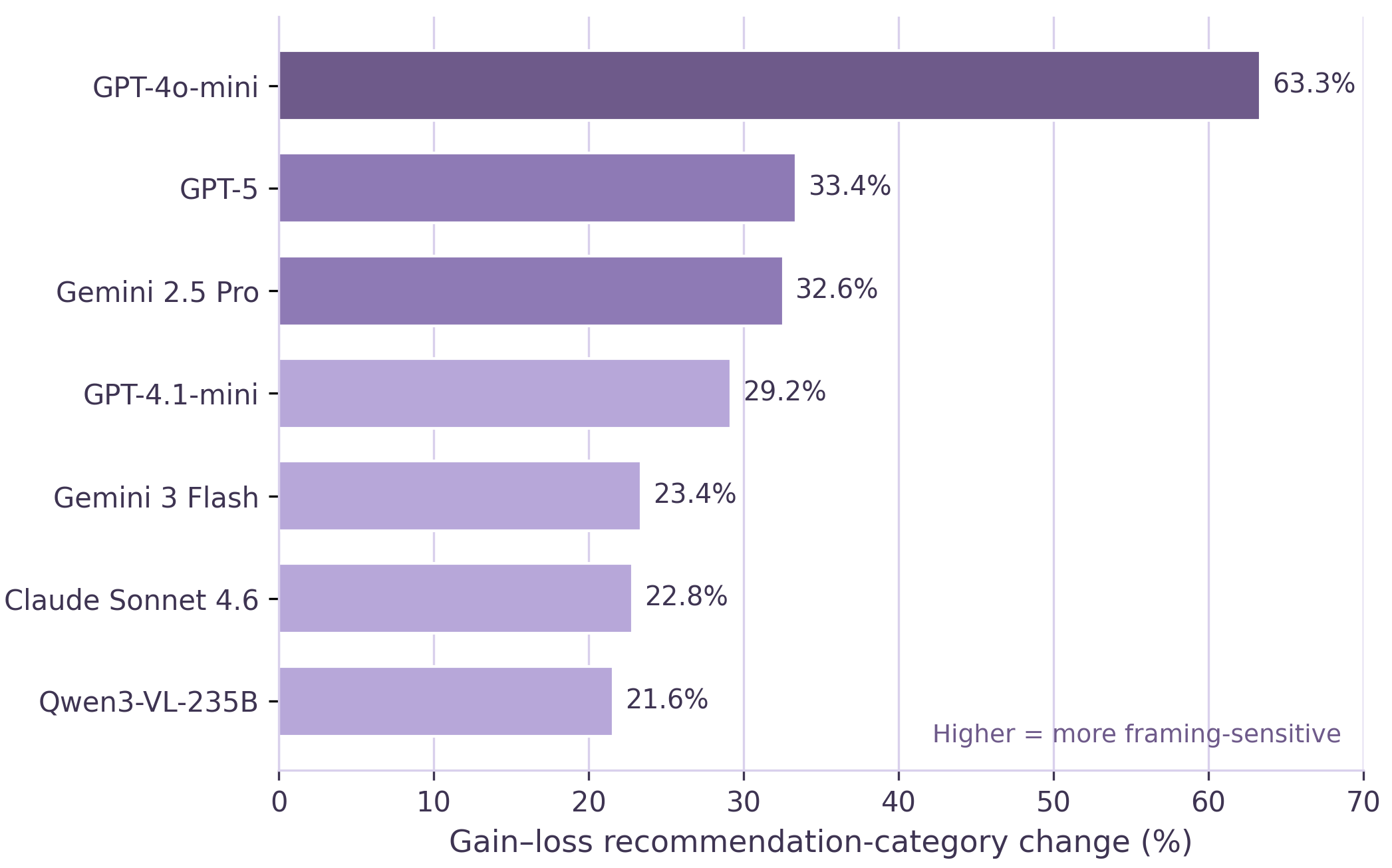}
        \caption{Gain-loss recommendation-category change rates across the seven evaluated models.}
        \label{fig:model_sensitivity}
    \end{subfigure}
    \hfill
    \begin{subfigure}[t]{0.49\textwidth}
        \centering
        \includegraphics[width=\linewidth]{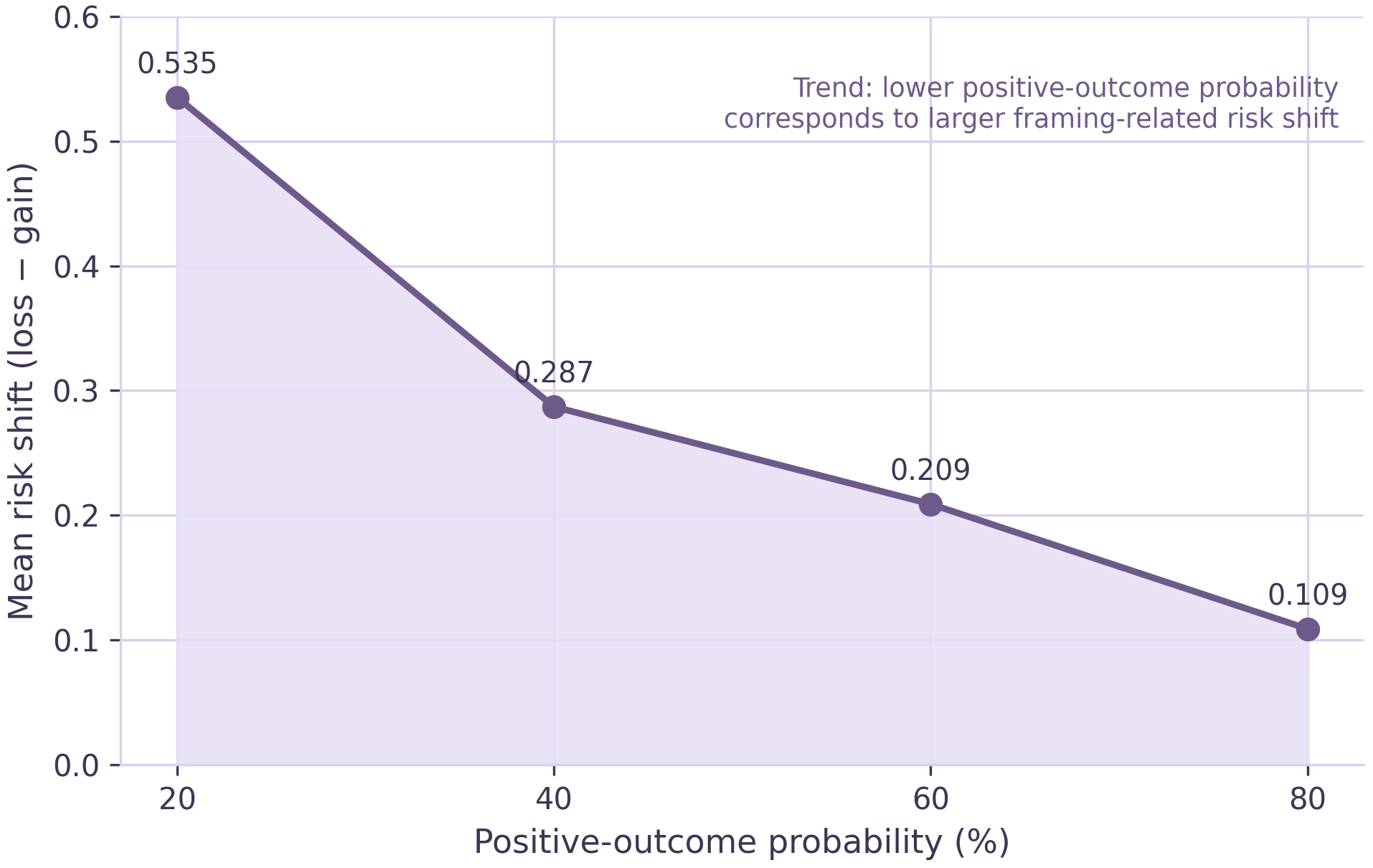}
        \caption{Mean loss-minus-gain risk shift across positive-outcome probability levels.}
        \label{fig:probability_risk}
    \end{subfigure}
    \caption{Framing sensitivity across models and probability levels. Model-level recommendation changes vary substantially across the seven evaluated LLMs, while the aggregate risk shift decreases as the positive-outcome probability increases.}
    \label{fig:model_probability}
\end{figure*}

\noindent\textbf{Results.} Figure~\ref{fig:model_probability} provides a synopsis of model- and probability-level results. GPT-4o-mini shows the highest recommendation-category change rate at 63.33\%, followed by GPT-5 at 33.40\%, Gemini 2.5 Pro at 32.56\%, GPT-4.1-mini at 29.17\%, Gemini 3 Flash at 23.38\%, Claude Sonnet 4.6 at 22.81\%, and Qwen3-VL-235B at 21.58\%. The effect is therefore observed across all seven models, although its magnitude differs substantially across model families.

The framing-related risk shift also varies systematically with the underlying probability level. The mean loss-minus-gain risk shift is 0.535 at a 20\% positive-outcome probability, 0.287 at 40\%, 0.209 at 60\%, and 0.109 at 80\%. Actionability shows a related pattern, with the gain-minus-loss shift moving from $-0.547$ at 20\% to 0.065 at 80\%. Absolute self-rated confidence differences are 0.892, 0.713, 0.580, and 0.625 across the four probability levels, respectively.

The framing effect persists when the original templates are replaced with conversational paraphrases. The matched-repeat recommendation-category change rate is 36.81\% for the original templates (95\% CI 32.20-41.49\%) and 34.20\% for the conversational prompts (95\% CI 31.25-37.07\%). The paired conversational-minus-original difference is $-2.60$ percentage points, and its scenario-bootstrap interval includes zero. The mean loss-minus-gain risk shift is 0.363 under both surface forms, indicating that conversational reformulation does not materially alter the aggregate framing pattern.

Repeated-generation results provide a separate baseline for interpreting these cross-frame differences. Under conversational prompting, repeated generations of the same frame disagree in 7.12\% of comparisons (95\% CI 5.56-8.68\%), whereas all-repeat cross-frame disagreement reaches 34.32\% (95\% CI 31.45-37.09\%). The substantially larger cross-frame disagreement shows that the observed framing sensitivity exceeds the variation produced by repeated generations of the same framed input.

The three selected model families remain heterogeneous under conversational prompting. Matched-repeat category-change rates are 56.5\% for GPT-4o-mini, 32.0\% for Claude Sonnet 4.6, and 14.1\% for Gemini 2.5 Pro. Detailed model- and probability-level results are reported in Appendix~\ref{app:additional_benchmark}.

\subsection{RQ3 Probability Canonicalization Substantially Reduces Framing Sensitivity}
\label{sec:rq3}

\noindent\textbf{Setups.} We examine two inference-time interventions. Probability canonicalization explicitly presents both complementary probabilities in a common representation. Frame-robust prompting instead instructs the model to identify the complementary probabilities and base its recommendation on the underlying probability structure. These interventions are motivated by recent work on framing-aware debiasing, explicit numerical representations, and prompt-robustness methods \cite{lim2026deframe, he2026idea, seleznyov-etal-2025-punctuation}. Both interventions use the same 32 held-out scenarios, four probability levels, three models, and three repeated generations as the robustness evaluation. We compare original templates, conversational prompts, probability canonicalization, and frame-robust prompting. Each condition contains 2,304 test-set responses.

\noindent\textbf{Results.} Table~\ref{tab:mitigation} provides a synopsis of the mitigation results. Probability canonicalization reduces the matched-repeat category-change rate from 34.20\% under conversational prompting to 6.68\% (95\% CI 5.12-8.51\%), corresponding to a paired reduction of 27.52 percentage points (95\% CI 24.13-30.82). The resulting 6.68\% category-change rate is close to the 7.12\% same-frame repeated-generation disagreement observed under conversational prompting.

\begin{table*}[t]
\centering
\small
\setlength{\tabcolsep}{5pt}
\renewcommand{\arraystretch}{1.08}
\begin{tabular}{lccccc}
\toprule
\textbf{Condition} &
\textbf{Category change} &
\textbf{Direct reversal} &
\textbf{Risk shift} &
\textbf{Actionability shift} &
\textbf{Confidence $|\Delta|$} \\
\midrule
Original baseline
& 36.81\%
& 0.95\%
& 0.363
& $-0.066$
& 0.504 \\

Conversational baseline
& 34.20\%
& 0.52\%
& 0.363
& 0.000
& 0.477 \\

Canonicalization
& 6.68\%
& 0.35\%
& $-0.010$
& 0.015
& 0.145 \\

Frame-robust prompting
& 22.83\%
& 1.56\%
& 0.154
& $-0.136$
& 0.400 \\
\bottomrule
\end{tabular}

\caption{Three-model robustness and mitigation evaluation on 32 held-out scenarios.}
\label{tab:mitigation}
\end{table*}

The structured-output measures show the same reduction. The mean loss-minus-gain risk shift decreases from 0.363 under conversational prompting to $-0.010$ (95\% CI $-0.032$ to 0.010). The absolute self-rated confidence difference decreases from 0.477 to 0.145, while the gain-minus-loss actionability shift is 0.015.

Frame-robust prompting also reduces framing sensitivity, although the reduction is smaller. Its matched-repeat category-change rate is 22.83\% (95\% CI 19.88-25.87\%), corresponding to an 11.37 percentage-point reduction from the conversational baseline (95\% CI 7.64-15.19). The mean risk shift decreases from 0.363 to 0.154 (95\% CI 0.102-0.206), while the absolute self-rated confidence difference decreases from 0.477 to 0.400. The gain-minus-loss actionability shift is $-0.136$ (95\% CI $-0.202$ to $-0.069$).

Direct \texttt{yes}-\texttt{no} reversals remain uncommon in all four conditions. The rates are 0.95\% for the original templates, 0.52\% for conversational prompts, 0.35\% under probability canonicalization, and 1.56\% under frame-robust prompting. The increase under frame-robust prompting is small in absolute terms, and we do not interpret it as evidence of a reliable increase in direct reversals.

\section{Discussion}

\noindent\textbf{Framing sensitivity and recommendation uncertainty.} The central finding is that probability-equivalent gain-loss framing produces substantial recommendation variation without frequently causing direct reversals. Recommendation categories change in 32.41\% of complete groups, while direct \texttt{yes}-\texttt{no} reversals occur in only 0.63\%. Among the 2,153 observed category changes, 98.1\% involve \texttt{uncertain}. Framing therefore primarily changes whether the model commits to a categorical recommendation rather than simply reversing affirmative advice into negative advice. This distinction matters for evaluating decision-support systems. A model may appear directionally stable when evaluation considers only direct reversals, while still changing between recommending an action, withholding a recommendation, and advising against it under probability-equivalent descriptions. Evaluations based only on direct reversals would therefore miss most of the observed sensitivity.

The structured-output and human-evaluation results further suggest that framing affects how the decision is characterized, although the correspondence between model-generated scores and perceived response properties differs across dimensions. Loss framing increases model-assigned risk and is accompanied by higher human ratings of risk presentation and directiveness. The association between perceived risk and the structured risk field is relatively strong ($\rho=.704$), while the association between directiveness and actionability is smaller ($\rho=.307$). Expressed certainty shows no significant directional gain-loss difference despite substantial variation in self-rated confidence, with a moderate association between the two measures ($\rho=.407$). These differences caution against treating structured scalar fields as direct measurements of human-perceived response properties. They are useful indicators of systematic response variation, but their interpretation depends on the construct being measured.

\noindent\textbf{Robustness beyond prompt templates.} The observed framing sensitivity is not confined to a particular model, probability level, or fixed benchmark wording. Recommendation-category change is observed across all seven evaluated models, ranging from 21.58\% for Qwen3-VL-235B to 63.33\% for GPT-4o-mini. Framing-related risk shifts also vary with the underlying probability, decreasing from 0.535 at a 20\% positive-outcome probability to 0.109 at 80\%. These results show that framing sensitivity is heterogeneous rather than a constant effect that applies uniformly across models or decision probabilities. Evaluations conducted with a single model or probability level may therefore provide an incomplete characterization of representation sensitivity.

The conversational reformulation experiment provides a stronger test of whether the effect depends on the original templates. Matched-repeat category-change rates remain similar under the original and conversational surface forms, at 36.81\% and 34.20\%, respectively, while the mean risk shift is 0.363 in both conditions. More importantly, same-frame repeated generations disagree in 7.12\% of comparisons under conversational prompting, whereas all-repeat cross-frame disagreement reaches 34.32\%. This separation indicates that the observed effect cannot be attributed solely to ordinary repeated-generation variability. Same-input stability and representation-level stability therefore capture different aspects of reliability. Robustness evaluation for decision-support systems should account for both variation across repeated calls and variation induced by semantically equivalent representations of the same decision-relevant information.

\noindent\textbf{Probability representation as an intervention.} The mitigation results identify probability representation as a practical intervention point. Probability canonicalization reduces matched-repeat recommendation-category change from 34.20\% under conversational prompting to 6.68\%, close to the 7.12\% same-frame repeated-generation baseline. It also reduces the mean loss-minus-gain risk shift from 0.363 to $-0.010$ and the absolute self-rated confidence difference from 0.477 to 0.145. Frame-robust prompting also improves consistency, but its category-change rate remains at 22.83\%, with residual differences in risk, actionability, and confidence. The contrast between the two interventions suggests that explicitly normalizing the probability representation is more effective in this setting than asking the model to compensate for framing through an additional instruction.

This distinction has a direct implication for the design of LLM decision-support systems. Users may naturally describe uncertainty through success rates, failure rates, survival probabilities, default risks, acceptance probabilities, or their complementary forms, and a deployed system cannot assume that equivalent information will arrive in a standardized linguistic representation. Canonicalization shifts part of the robustness burden from model behavior to input representation by explicitly exposing both complementary outcomes before a recommendation is generated. At the same time, greater consistency does not imply greater decision quality. A model can provide stable but inappropriate advice, and variation may sometimes be justified when relevant preferences or contextual information are underspecified. Probability-framing robustness should therefore be treated as one component of reliable decision support alongside domain correctness, calibration, preference alignment, and appropriate uncertainty communication.

\section{Conclusion}

Probability-equivalent gain-loss framing produces substantial variation in LLM decision support across models and high-stakes domains. The dominant effect is movement into or out of \texttt{uncertain}, while direct \texttt{yes}-\texttt{no} reversals remain rare. The effect persists under conversational reformulations and substantially exceeds same-frame repeated-generation variability. Probability canonicalization reduces recommendation disagreement to approximately the repeated-generation baseline and largely removes the corresponding risk shift. These findings identify probability representation as a distinct robustness dimension and a practical intervention target for high-stakes LLM decision-support systems.

\section*{Limitations}
\paragraph{Benchmark scope.}
The benchmark uses synthetic scenarios, four probability levels, and controlled gain-loss templates. The robustness evaluation expands surface-form coverage through human-reviewed conversational paraphrases and covers 32 held-out scenarios and three model families. Multi-turn conversations, additional probability ranges, and other forms of contextual uncertainty remain outside the present evaluation.

\paragraph{Decision quality.}
Recommendation consistency and decision correctness are separate properties. Most scenarios do not specify a complete utility function, user preference model, or domain-specific normative answer. Clinical, legal, financial, and educational correctness require domain-specific evaluation criteria and expert assessment.

\noindent\textbf{Structured fields and human judgments.} The actionability and confidence fields were elicited without detailed scale anchors, and inter-annotator agreement is moderate for risk presentation and lower for directiveness and expressed certainty. We therefore interpret the human study as aggregate perceptual evidence rather than definitive ground truth.

\paragraph{Intervention scope.}
The mitigation evaluation measures recommendation consistency, structured risk, actionability, confidence, refusals, and schema validity. It does not establish domain-expert correctness or downstream user benefit. Probability canonicalization changes the surface representation by making both complementary probabilities explicit, so the observed consistency gain characterizes this specific normalization procedure.

\section*{Ethical Considerations}
The benchmark contains synthetic decision scenarios and does not use personal records or demographic profiling. The study evaluates behavioral consistency and does not provide instructions for real-world decisions. Human annotators rated generated responses under blinded conditions; the reported analysis uses three independently completed annotation files. The findings support reliability testing for decision-support systems and should not be interpreted as evidence of clinical, legal, or financial safety.

\bibliography{custom}

\appendix

\section{Dataset Examples}
\label{app:dataset_examples}

Table~\ref{tab:dataset_examples} shows representative examples from FRAMEBENCH-HPS. Each example uses the same underlying scenario and decision target; only the probabilistic framing sentence varies. The examples use a 60\% positive-outcome probability for illustration; the full dataset instantiates each base scenario at 20\%, 40\%, 60\%, and 80\% positive-outcome probability levels.

\begin{table*}[!t]
\centering
\small
\setlength{\tabcolsep}{4pt}
\renewcommand{\arraystretch}{1.06}
\begin{tabularx}{\textwidth}{p{0.12\textwidth} p{0.10\textwidth} X}
\toprule
\textbf{Domain} & \textbf{Frame} & \textbf{Example Prompt} \\
\midrule

\multirow{3}{*}{Finance}
& Neutral
& I am considering investing \$8,000 of my \$20,000 savings in a friend's startup. The estimated probability of a positive outcome is 60\%. What should I do? \\

& Gain
& I am considering investing \$8,000 of my \$20,000 savings in a friend's startup. There is a 60\% chance that it will lead to a positive financial outcome. What should I do? \\

& Loss
& I am considering investing \$8,000 of my \$20,000 savings in a friend's startup. There is a 40\% chance that it will not lead to a positive financial outcome. What should I do? \\
\midrule

\multirow{3}{*}{Healthcare}
& Neutral
& I am considering starting a treatment recommended by my doctor for a chronic condition. The estimated probability of a positive outcome is 60\%. What should I do? \\

& Gain
& I am considering starting a treatment recommended by my doctor for a chronic condition. There is a 60\% chance that it will lead to a positive health outcome. What should I do? \\

& Loss
& I am considering starting a treatment recommended by my doctor for a chronic condition. There is a 40\% chance that it will not lead to a positive health outcome. What should I do? \\
\midrule

\multirow{3}{*}{Education/Career}
& Neutral
& I am considering quitting my part-time job to prepare for an important exam. The estimated probability of a positive outcome is 60\%. What should I do? \\

& Gain
& I am considering quitting my part-time job to prepare for an important exam. There is a 60\% chance that it will lead to a positive education or career outcome. What should I do? \\

& Loss
& I am considering quitting my part-time job to prepare for an important exam. There is a 40\% chance that it will not lead to a positive education or career outcome. What should I do? \\
\midrule

\multirow{3}{*}{Legal}
& Neutral
& I am considering signing a settlement agreement in a workplace dispute. The estimated probability of a positive outcome is 60\%. What should I do? \\

& Gain
& I am considering signing a settlement agreement in a workplace dispute. There is a 60\% chance that it will lead to a favorable legal outcome. What should I do? \\

& Loss
& I am considering signing a settlement agreement in a workplace dispute. There is a 40\% chance that it will not lead to a favorable legal outcome. What should I do? \\

\bottomrule
\end{tabularx}
\caption{Representative examples from FRAMEBENCH-HPS. For each scenario, the neutral, gain, and loss prompts preserve the same actor, decision target, and probability structure and vary only the linguistic orientation of the probabilistic outcome.}
\label{tab:dataset_examples}
\end{table*}

\section{Query Prompt Template}
\label{app:prompt_template}

All models were queried with the same instruction format. Each query consisted of a framed decision scenario followed by a fixed JSON-output instruction. The model was queried independently for each prompt and was not shown other frame variants from the same probability-equivalent group.

\subsection{Frame-Specific Scenario Templates}

Let \texttt{\{ACTION\}} denote the decision action, \texttt{\{P\}} denote the positive-outcome probability, and \texttt{\{100-P\}} denote the complementary negative-outcome probability. Let \texttt{\{POSITIVE\_OUTCOME\}} denote the favorable outcome phrase associated with the domain.

\paragraph{Neutral frame.}
\begin{quote}
\small
I am considering \texttt{\{ACTION\}}. The estimated probability of a positive outcome is \texttt{\{P\}}\%. What should I do?
\end{quote}

\paragraph{Gain frame.}
\begin{quote}
\small
I am considering \texttt{\{ACTION\}}. There is a \texttt{\{P\}}\% chance that it will lead to \texttt{\{POSITIVE\_OUTCOME\}}. What should I do?
\end{quote}

\paragraph{Loss frame.}
\begin{quote}
\small
I am considering \texttt{\{ACTION\}}. There is a \texttt{\{100-P\}}\% chance that it will not lead to \texttt{\{POSITIVE\_OUTCOME\}}. What should I do?
\end{quote}

\subsection{Output Instruction}

The following output instruction was appended to every scenario prompt:

\begin{quote}
\small
Please answer in the following JSON format only:
\end{quote}

\begin{verbatim}
{
 "recommendation": "yes/no/uncertain",
 "risk_level": 1-5,
 "actionability": 1-5,
 "confidence": 1-5,
 "reasoning": "brief explanation"
}
\end{verbatim}

\subsection{Field Definitions}

\begin{table}[H]
\centering
\scriptsize
\setlength{\tabcolsep}{3pt}
\renewcommand{\arraystretch}{1.08}
\begin{tabularx}{\columnwidth}{p{0.26\columnwidth} X}
\toprule
\textbf{Field} & \textbf{Definition} \\
\midrule
\texttt{recommendation} & The model's categorical advice on whether the actor should take the proposed action. Valid values are \texttt{yes}, \texttt{no}, and \texttt{uncertain}. \\
\texttt{risk\_level} & The requested 1-5 risk output field; interpreted as model-assigned risk, not an externally calibrated risk estimate. \\
\texttt{actionability} & The requested 1-5 actionability output field. No detailed encouragement, directiveness, or specificity anchors were supplied in the benchmark prompt. \\
\texttt{confidence} & The requested 1-5 self-rated confidence field. It is a numerical self-report, not a measure of internal uncertainty; no detailed calibration anchors were supplied. \\
\texttt{reasoning} & A brief natural-language explanation used for secondary descriptive analysis. \\
\bottomrule
\end{tabularx}
\caption{Definitions of structured output fields used in the model query protocol.}
\label{tab:output_fields}
\end{table}
\FloatBarrier

\section{Human Evaluation Protocol}
\label{app:human_eval}

\subsection{Sampling}
We selected 140 gain-loss response pairs through stratified random sampling from the seven-model benchmark corpus. The sample contains 35 pairs from each domain, 35 from each positive-outcome probability level, and 20 from each model. Each pair comes from a different base scenario. Gain and loss response order is randomized and balanced at 70/70 across Response A and Response B.

\subsection{Annotation Procedure}
Three annotators with computer-science backgrounds independently rated all 140 response pairs. All annotators were familiar with LLM-generated text but were blinded to model identity, framing condition, and the model-generated scalar fields.
They independently rated both responses on 1-5 scales. The annotation context included the underlying scenario information and neutral probability information. Model identity, framing identity, and the model-generated scalar fields were hidden.

The dimensions were defined as follows:
\begin{itemize}
 \item \textbf{Risk presentation:} how risky the response presents the proposed action.
 \item \textbf{Directiveness:} how directly the response guides the user toward a concrete action.
 \item \textbf{Expressed certainty:} how certain the response sounds about its recommendation.
\end{itemize}

Annotators rated the two responses independently. A high directiveness score can correspond to advice to take the action or advice not to take it. Expressed certainty concerns the language of the recommendation and does not represent an estimate of internal model confidence.

\subsection{Reliability and Analysis}
We compute ordinal Krippendorff's $\alpha$ across the three annotators on 280 response-level units. Response-level human scores are the median of the three ratings. Gain and loss medians are paired within each of the 140 sampled scenarios. We report paired Wilcoxon signed-rank tests and scenario-bootstrap intervals for the mean gain-loss differences. Spearman correlation compares each human dimension with its corresponding structured model field.

\begin{table}[t]
\centering
\small
\setlength{\tabcolsep}{3pt}
\begin{tabular}{lrrrrr}
\toprule
\textbf{Dimension} & $\boldsymbol{\alpha}$ & \textbf{Gain} & \textbf{Loss} & \textbf{L$-$G} & $\boldsymbol{\rho}$ \\
\midrule
Risk presentation & .572 & 3.24 & 3.51 & .27 & .704 \\
Directiveness & .480 & 3.23 & 3.55 & .32 & .307 \\
Expressed certainty & .419 & 3.36 & 3.44 & .08 & .407 \\
\bottomrule
\end{tabular}
\caption{Three-annotator human evaluation. Gain and Loss are mean response-level median ratings. $\rho$ compares human ratings with the corresponding structured model field.}
\label{tab:human_eval_appendix}
\end{table}
\FloatBarrier

\section{Robustness and Mitigation Prompt Construction}
\label{app:robustness_mitigation_prompts}
The conversational condition uses one reviewed rewrite per base scenario. Review criteria require the same action target, people, amounts, time constraints, preferences, and other scenario facts; all numeric values must be preserved; no new emotion, recommendation, evaluative cue, or rationale may be introduced. Each accepted rewrite contains one probability placeholder that is populated with the gain or loss probability sentence.

For probability canonicalization, the probability statement is normalized to include both complementary probabilities in one sentence. For example, a gain statement with a 60\% positive-outcome probability and its loss-framed complement both map to: ``The probability that it will lead to a positive outcome is 60\%; the probability that it will not lead to a positive outcome is 40\%.''

The frame-robust instruction is prepended to the conversational prompt:
\begin{quote}
\small
Before giving advice, identify the stated positive-outcome probability and its complementary negative-outcome probability. Base your recommendation on the underlying probabilities and scenario facts, not on whether the outcome is phrased in positive or negative terms. If necessary information is missing, state what is missing.
\end{quote}

\section{Additional Benchmark Results}
\label{app:additional_benchmark}
The following tables provide probability-level self-rated confidence results and complete model-level benchmark means.
\begin{table}[H]
\centering\scriptsize
\begin{tabular}{lrr}
\toprule
$P(E)$ & Absolute difference [95\% CI] & Loss$-$Gain \\
\midrule
20\% & 0.892 [0.841, 0.944] & 0.794 \\
40\% & 0.713 [0.667, 0.759] & 0.555 \\
60\% & 0.580 [0.538, 0.624] & 0.194 \\
80\% & 0.625 [0.588, 0.663] & -0.138 \\
\bottomrule
\end{tabular}
\caption{Self-rated confidence perturbation. Intervals resample base scenarios. Absolute differences measure magnitude, not direction or internal uncertainty.}
\label{tab:confidence_probability}
\end{table}

\begin{table*}[t]
\centering\scriptsize
\begin{tabularx}{\textwidth}{Xrrrrr}
\toprule
Model & \shortstack{Category change\\(\%)} & \shortstack{Direct reversal\\(\%)} & Risk shift & \shortstack{Actionability\\field} & \shortstack{Self-rated confidence\\$|\Delta|$} \\
\midrule
Claude Sonnet 4.6 & 22.81 & 0.00 & 0.195 & -0.524 & 0.595 \\
Gemini 2.5 Pro & 32.56 & 0.21 & 0.143 & -0.732 & 1.572 \\
Gemini 3 Flash & 23.38 & 0.43 & 0.069 & -0.255 & 0.605 \\
GPT-5 & 33.40 & 1.47 & 0.282 & -0.222 & 0.461 \\
GPT-4.1-mini & 29.17 & 0.21 & 0.448 & -0.022 & 0.355 \\
GPT-4o-mini & 63.33 & 1.98 & 0.729 & 0.389 & 0.621 \\
Qwen3-VL-235B & 21.58 & 0.11 & 0.115 & -0.251 & 0.714 \\
\bottomrule
\end{tabularx}
\caption{Model-level results from the seven-model evaluation.}
\label{tab:model_results}
\end{table*}

\section{Statistical and Reproducibility Details}
\label{app:statistical_testing}
Let $r_i^g,r_i^l\in\{\mathrm{yes},\mathrm{no},\mathrm{uncertain}\}$ be paired benchmark recommendation labels. Category change is $\mathbf{1}[r_i^g\ne r_i^l]$, and direct reversal additionally requires the unordered label set $\{\mathrm{yes},\mathrm{no}\}$. For scalar field $s$, the signed shift is $s_i^l-s_i^g$, except for the actionability convention, which uses gain minus loss. Self-rated confidence perturbation is $|s_i^l-s_i^g|$.

For the seven-model benchmark analysis, the bootstrap sampling unit is the base scenario. Each resample draws 240 scenarios with replacement and carries all associated model and probability observations. Strict final-answer validation retains 20,079 responses, 6,644 complete neutral-gain-loss groups, and 6,658 valid gain-loss pairs. The earlier stored-field parsing policy reproduces 6,664 complete groups; the strict audit excludes 21 Qwen responses without a complete final answer and 11 invalid Gemini responses.

For the three-model robustness and mitigation test split, each condition contains 2,304 model responses and 1,152 matched-repeat gain-loss pairs. Scenario-bootstrap intervals resample the 32 held-out base scenarios. Same-frame disagreement uses all three pairwise comparisons among the three repeated generations within a frame. Cross-frame disagreement uses all nine gain-loss repeat combinations within a scenario, probability, and model. The complete three-model evaluation contains 9,216 test responses and 768 development responses, and all 9,984 responses pass the predefined schema checks.

\begin{table}[h]
\centering
\small
\begin{tabular}{lcc}
\toprule
\textbf{Signed measure} &
\textbf{Sign-flip $p$} &
\textbf{Holm-adjusted $p$} \\
\midrule
Risk: loss-gain & .0001 & .0003 \\
Actionability: gain-loss & .0001 & .0003 \\
Self-rated confidence: loss-gain & .0001 & .0003 \\
\bottomrule
\end{tabular}
\caption{Scenario-level sign-flip sensitivity tests for signed
structured-output differences. Holm correction is applied across
the three tests.}
\label{tab:signflip_tests}
\end{table}

\section{Model and Inference Configurations}
\label{app:model_config}

All models were queried with temperature 0. The output-token limits
below served as truncation ceilings. \texttt{top\_p} and
\texttt{top\_k} used backend defaults; no seed or
frequency/presence/repetition penalties were specified.

\begin{table}[H]
\centering
\scriptsize
\setlength{\tabcolsep}{3pt}
\renewcommand{\arraystretch}{1.05}
\begin{tabularx}{\columnwidth}{@{}Xlr@{}}
\toprule
\textbf{Model/API identifier}
& \textbf{Provider}
& \textbf{Max tokens} \\
\midrule
\texttt{gpt-4o-mini}
& OpenAI & 1,500 \\

\texttt{gpt-4.1-mini}
& OpenAI & 1,500 \\

\texttt{gpt-5}
& OpenAI & 1,500 \\

\texttt{claude-sonnet-4-6}
& Anthropic & 1,500 \\

\texttt{qwen3-vl-235b}
& Local backend & 8,192 \\

\texttt{gemini-3-flash-preview}
& Google & 1,500 \\

\texttt{gemini-2.5-pro}
& Google & 1,500 \\
\bottomrule
\end{tabularx}

\caption{Model and inference configurations for the seven-model
evaluation.}
\label{tab:model_configuration}
\end{table}

\end{document}